\documentclass[10pt,twocolumn,letterpaper]{article}

\usepackage{cvpr}
\usepackage{times}
\usepackage{epsfig}
\usepackage{graphicx}
\usepackage{amsmath}
\usepackage{amssymb}

\usepackage[breaklinks=true,bookmarks=false]{hyperref}

\cvprfinalcopy 

\def\cvprPaperID{618} 

\begin{document}

\title{NTU RGB+D: A Large Scale Dataset for 3D Human Activity Analysis}

\author{Amir  Shahroudy$^{\dagger,\ddagger}$\\{\tt\small amir3@ntu.edu.sg}
	\and Jun  Liu$^{\dagger}$\\{\tt\small jliu029@ntu.edu.sg}
	\and Tian-Tsong  Ng$^{\ddagger}$\\{\tt\small ttng@i2r.a-star.edu.sg}
	\and Gang  Wang$^{\dagger,}$\thanks{Corresponding author} \\{\tt\small wanggang@ntu.edu.sg}
	\and $\dagger$ School of Electrical and Electronic Engineering, Nanyang Technological University, Singapore
	\\$\ddagger$ Institute for Infocomm Research, Singapore
}
\maketitle

\begin{abstract}
Recent approaches in depth-based human activity analysis achieved outstanding performance and proved the effectiveness of 3D representation for classification of action classes.
Currently available depth-based and RGB+D-based action recognition benchmarks have a number of limitations, including the lack of training samples, distinct class labels, camera views and variety of subjects.
In this paper we introduce a large-scale dataset for RGB+D human action recognition with more than 56 thousand video samples and 4 million frames, collected from 40 distinct subjects.
Our dataset contains 60 different action classes including daily, mutual, and health-related actions.
In addition, we propose a new recurrent neural network structure to model the long-term temporal correlation of the features for each body part, and utilize them for better action classification.
Experimental results show the advantages of applying deep learning methods over state-of-the-art hand-crafted features on the suggested cross-subject and cross-view evaluation criteria for our dataset.
The introduction of this large scale dataset will enable the community to apply, develop and adapt various data-hungry learning techniques for the task of depth-based and RGB+D-based human activity analysis.
\end{abstract}

\begin{table*}
	\setlength{\tabcolsep}{3pt} 
	\begin{center}
		\begin{tabular}{|lr|r|r|r|c|l|c|c|}
			\hline
			Datasets			& ~	 						&Samples	&Classes&Subjects	&Views	&Sensor 		& Modalities			& Year	\\\hline\hline
			MSR-Action3D		& \cite{msraction3ddataset}	& 567		& 20	& 10		& 1		& N/A		& D+3DJoints			& 2010	\\\hline
			CAD-60				& \cite{cad60}				& 60		& 12	& 4			& -		& Kinect v1 	& RGB+D+3DJoints		& 2011	\\\hline
			RGBD-HuDaAct		& \cite{rgbdhudaact}		& 1189		& 13	& 30		& 1		& Kinect v1		& RGB+D					& 2011	\\\hline
			MSRDailyActivity3D	& \cite{actionletCVPR}		& 320		& 16	& 10		& 1		& Kinect v1		& RGB+D+3DJoints		& 2012	\\\hline
			Act$4^2$			& \cite{Act42}				& 6844		& 14	& 24		& 4		& Kinect v1		& RGB+D					& 2012	\\\hline
			CAD-120				& \cite{cad120}				& 120		& 10+10	& 4			& -		& Kinect v1		& RGB+D+3DJoints		& 2013	\\\hline
			3D Action Pairs		& \cite{HON4D}				& 360		& 12	& 10		& 1		& Kinect v1		& RGB+D+3DJoints		& 2013	\\\hline
			Multiview 3D Event	& \cite{Multiview3DEvent}	& 3815		& 8		& 8			& 3		& Kinect v1		& RGB+D+3DJoints		& 2013	\\\hline
			Online RGB+D Action	& \cite{Orderlet}			& 336		&	7	& 24		& 1		& Kinect v1		& RGB+D+3DJoints		& 2014	\\\hline
			Northwestern-UCLA	& \cite{NW_UCLA}			& 1475		& 10	& 10		& 3		& Kinect v1		& RGB+D+3DJoints		& 2014	\\\hline
			UWA3D Multiview 	& \cite{HOPC}				&$\sim$900	& 30	& 10		& 1		& Kinect v1		& RGB+D+3DJoints		& 2014	\\\hline
			Office Activity		& \cite{OfficeActivity}		& 1180		& 20	& 10		& 3		& Kinect v1		& RGB+D					& 2014	\\\hline
			UTD-MHAD			& \cite{UTD-MHAD}			& 861		& 27	& 8			& 1		& Kinect v1+WIS	& RGB+D+3DJoints+ID		& 2015	\\\hline
			UWA3D Multiview II	& \cite{HOPC_PAMI}			& 1075		& 30	& 10		& 5		& Kinect v1		& RGB+D+3DJoints		& 2015	\\\hline
			\hline
			\bf NTU RGB+D 		& ~ 						& \bf 56880		& \bf 60	& \bf 40 		& \bf 80	& \bf Kinect v2		& \bf RGB+D+IR+3DJoints 	& \bf 2016	\\\hline
		\end{tabular}
	\end{center}
	\caption{Comparison between NTU RGB+D dataset and some of the other publicly available datasets for 3D action recognition.
		Our dataset provides many more samples, action classes, human subjects, and camera views in comparison with other available datasets for RGB+D action recogniton.}
	\label{tab:datasetcomparison}
\end{table*}

\section{Introduction}
\label{sec:intro}


Recent development of depth sensors enabled us to obtain effective 3D structures of the scenes and objects \cite{kinectSurvey2013}.
This empowers the vision solutions to move one important step towards 3D vision, \eg 3D object recognition, 3D scene understanding, and 3D action recognition \cite{aggarwal2014survey}.

Unlike the RGB-based counterpart, 3D video analysis suffers from the lack of large-sized benchmark datasets.
Yet there are no any sources of publicly shared 3D videos such as YouTube to supply ``in-the-wild'' samples.
This limits our ability to build large-sized benchmarks to evaluate and compare the strengths of different methods, especially the recent data-hungry techniques like deep learning approaches.
To the best of our knowledge, all the current 3D action recognition benchmarks have limitations in various aspects.

First is the small number of subjects and very narrow range of performers' ages, which makes the intra-class variation of the actions very limited.
The constitution of human activities depends on the age, gender, culture and even physical conditions of the subjects.
Therefore, variation of human subjects is crucial for an action recognition benchmark.

Second factor is the number of the action classes.
When only a very small number of classes are available, each action class can be easily distinguishable by finding a simple motion pattern or even the appearance of an interacted object.
But when the number of classes grows, the motion patterns and interacting objects will be shared between classes and the classification task will be more challenging.

Third is the highly restricted camera views. 
For most of the datasets, all the samples are captured from a front view with a fixed camera viewpoint. 
For some others, views are bounded to fixed front and side views, using multiple cameras at the same time. 

Finally and most importantly, the highly limited number of video samples prevents us from applying the most advanced data-driven learning methods to this problem. 
Although some attempts have been done \cite{rnnskeleton_cvpr15,cnn_for_depth_action_THMS}, they suffered from overfitting and had to scale down the size of learning parameters; 
as a result, they clearly need many more samples to generalize and perform better on testing data.

To overcome these limitations, we develop a new large-scale benchmark dataset for 3D human activity analysis. 
The proposed dataset consists of $56,880$ RGB+D video samples, captured from 40 different human subjects, using Microsoft Kinect v2.
We have collected RGB videos, depth sequences, skeleton data (3D locations of 25 major body joints), and infrared frames.
Samples are captured in 80 distinct camera viewpoints.
The age range of the subjects in our dataset is from 10 to 35 years which brings more realistic variation to the quality of actions.
Although our dataset is limited to indoor scenes, due to the operational limitation of the acquisition sensor, we provide the ambiance inconstancy by capturing in various background conditions.
This large amount of variation in subjects and views makes it possible to have more accurate cross-subject and cross-view evaluations for various 3D-based action analysis methods.

The proposed dataset can help the community to move steps forward in 3D human activity analysis and makes it possible to apply data-hungry methods such as deep learning techniques for this task.

As another contribution, inspired by the physical characteristics of human body motion, we propose a novel part-aware extension of the long short-term memory (LSTM) model \cite{lstm}.
Human actions can be interpreted as interactions of different parts of the body.
In this way, the joints of each body part always move together and the combination of their 3D trajectories form more complex motion patterns.
By splitting the memory cell of the LSTM into part-based sub-cells, the recurrent network will learn the long-term patterns specifically for each body part and the output of the unit will be learned from the combination of all the sub-cells.

Our experimental results on the proposed dataset shows the clear advantages of data-driven learning methods over state-of-the-art hand-crafted features.

The rest of this paper is organized as follows:
Section \ref{sec:relatedwork} explores the current 3D-based human action recognition methods and benchmarks.
Section \ref{sec:dataset} introduces the proposed dataset, its structure, and defined evaluation criteria.
Section \ref{sec:approach} presents our new part-aware long short-term memory network for action analysis in a recurrent neural network fashion.
Section \ref{sec:exp} shows the experimental evaluations of state-of-the-art hand-crafted features alongside the proposed recurrent learning method on our benchmark,
and section \ref{sec:conclusion} concludes the paper.

\section{Related work}
\label{sec:relatedwork}

In this section we briefly review publicly available 3D activity analysis benchmark datasets and recent methods in this domain.
Here we introduce a limited number of the most famous ones.
For a more extensive list of current 3D activity analysis datasets and methods, readers are referred to these survey papers \cite{RGB-D_Survey_40_Pichao, aggarwal2014survey,Chen20131995,2016arXiv160101006H,IJPRAI_Survey_Kinect, surveyMaoYe, Cai2016Survey}.

\subsection{3D activity analysis datasets}
After the release of Microsoft Kinect \cite{kinectSurvey2012}, several datasets are collected by different groups to perform research on 3D action recognition and to evaluate different methods in this field.

MSR-Action3D dataset \cite{msraction3ddataset} was one of the earliest ones which opened up the research in depth-based action analysis.
The samples of this dataset were limited to depth sequences of gaming actions \eg \emph{forward punch, side-boxing, forward kick, side kick, tennis swing, tennis serve, golf swing, etc.}
Later the body joint data was added to the dataset.
Joint information includes the 3D locations of 20 different body joints in each frame.
A decent number of methods are evaluated on this benchmark and recent ones reported close to saturation accuracies \cite{Luo_2013_ICCV,RangeSample,MMMP_PAMI}.

CAD-60 \cite{cad60} and CAD-120 \cite{cad120} contain RGB, depth, and skeleton data of human actions.
The special characteristic of these datasets is the variety of camera views.
Unlike most of the other datasets, camera is not bound to front-view or side-views.
However, the limited number of video samples (60 and 120) is the downside of them.

RGBD-HuDaAct \cite{rgbdhudaact} was one of the largest datasets.
It contains RGB and depth sequences of 1189 videos of 12 human daily actions (plus one background class), with high variation in time lengths.
The special characteristic of this dataset was the synced and aligned RGB and depth channels which enabled local multimodal analysis of RBGD signals\footnote{
We emphasize the difference between RGBD and RGB+D terms.
We suggest to use RGBD when the two modalities are aligned pixel-wise, and RGB+D when the resolutions of the two are different and frames are not aligned.}.

MSR-DailyActivity \cite{actionletCVPR} was among the most challenging benchmarks in this field.
It contains 320 samples of 16 daily activities with higher intra-class variation.
Small number of samples and the fixed viewpoint of the camera are the limitations of this dataset.
Recently reported results on this dataset also achieved very high accuracies \cite{RangeSample, jianfang_CVPR15, Luo_2013_ICCV,DSSCA-arXiv}.

3D Action Pairs \cite{HON4D} was proposed to provide multiple pairs of action classes.
Each pair contains very closely related actions with differences along temporal axis \eg \emph{pick up/put down a box, push/pull a chair, wear/take off a hat, etc.}
State-of-the-art methods \cite{Kong_2015_CVPR,MMMP_PAMI,DSSCA-arXiv} achieved perfect accuracy on this benchmark.

Multiview 3D event \cite{Multiview3DEvent} and Northwestern-UCLA \cite{NW_UCLA} datasets used more than one Kincect cameras at the same time to collect multi-view representations of the same action, and scale up the number of samples.

It is worth mentioning, there are more than 40 datasets specifically for 3D human action recognition \cite{RGB-D_Survey_40_Pichao}. 
Although each of them provided important challenges of human activity analysis, they have limitations in some aspects.
\tablename{~\ref{tab:datasetcomparison}} shows the comparison between some of the current datasets with our large-scale RGB+D action recognition dataset.

To summarize the advantages of our dataset over the existing ones, NTU RGB+D has:
1- many more action classes,
2- many more samples for each action class,
3- much more intra-class variations (poses, environmental conditions, interacted objects, age of actors, ...),
4- more camera views,
5- more camera-to-subject distances, and
6- used Kinect v.2 which provides more accurate depth-maps and 3D joints, especially in a multi-camera setup compared to the previous version of Kinect.

\subsection{3D action recognition methods}

After the introduction of first few benchmarks, a decent number of methods were proposed and evaluated on them.

Oreifej \etal \cite{HON4D} calculated the four-dimensional normals (X-Y-depth-time) from depth sequences and accumulates them on spatio-temporal cubes as quantized histograms over 120 vertices of a regular polychoron.
The work of \cite{HOPC_PAMI} proposed histograms of oriented principle components of depth cloud points, in order to extract robust features against viewpoint variations.
Lu \etal \cite{RangeSample} applied $\tau$ test based binary range-sample features on depth maps and achieved robust representation against noise, scaling, camera views, and background clutter.
Yang and Tian \cite{Yang_2014_CVPR} proposed supernormal vectors as aggregated dictionary-based codewords of four-dimensional normals over space-time grids.

To have a view-invariant representation of the actions, features can be extracted from the 3D body joint positions which are available for each frame.
Evangelidis \etal \cite{skeletalQuads} divided the body into part-based joint quadruples and encodes the configuration of each part with a succinct 6D feature vector, so called skeletal quads.
To aggregate the skeletal quads, they applied Fisher vectors and classified the samples by a linear SVM.
In \cite{VemulapalliCVPR14} different skeleton configurations were represented as points on a Lie group.
Actions as time-series of skeletal configurations, were encoded as curves on this manifold.
The work of \cite{Luo_2013_ICCV} utilized group sparsity based class-specific dictionary coding with geometric constraints to extract skeleton-based features.
Rahmani and Mian \cite{Rahmani_2015_CVPR} introduced a nonlinear knowledge transfer model to transform different views of human actions to a canonical view.
To apply ConvNet-based learning to this domain, \cite{Rahmani_2016_CVPR} used synthetically generated data and fitted them to real mocap data.
Their learning method was able to recognize actions from novel poses and viewpoints.

In most of 3D action recognition scenarios, there are more than one modality of information and combining them helps to improve the classification accuracy.
Ohn-Bar and Trivedi \cite{hog2-ohnbar} combined second order joint-angle similarity representations of skeletons with a modified two step HOG feature on spatio-temporal depth maps to build global representation of each video sample and utilized a linear SVM to classify the actions.
Wang \etal \cite{actionletPAMI}, combined Fourier temporal pyramids of skeletal information with local occupancy pattern features extracted from depth maps and applied a data mining framework to discover the most discriminative combinations of body joints.
A structured sparsity based multimodal feature fusion technique was introduced by \cite{AmirAthens} for action recognition in RGB+D domain.
In \cite{6836044} random decision forests were utilized for learning and feature pruning over a combination of depth and skeleton-based features.
The work of \cite{MMMP_PAMI} proposed hierarchical mixed norms to fuse different features and select most informative body parts in a joint learning framework.
Hu \etal \cite{jianfang_CVPR15} proposed dynamic skeletons as Fourier temporal pyramids of spline-based interpolated skeleton points and their gradients, and HOG-based dynamic color and depth patterns to be used in a RGB+D joint-learning model for action classification.

{\bf RNN based 3D action recognition}:
The applications of recurrent neural networks for 3D human action recognition were explored very recently \cite{diffRNN, rnnskeleton_cvpr15, cooccurrance}.

Differential RNN \cite{diffRNN} added a new gating mechanism to the traditional LSTM to extract the derivatives of internal state (DoS).
The derived DoS was fed to the LSTM gates to learn salient dynamic patterns in 3D skeleton data.

HBRNN-L \cite{rnnskeleton_cvpr15} proposed a multilayer RNN framework for action recognition on a hierarchy of skeleton-based inputs. 
At the first layer, each subnetwork received the inputs from one body part.
On next layers, the combined hidden representation of previous layers were fed as inputs in a hierarchical combination of body parts.

The work of \cite{cooccurrance} introduced an internal dropout mechanism applied to LSTM gates for stronger regularization in the RNN-based 3D action learning network.
To further regularize the learning, a co-occurrence inducing norm was added to the network's cost function which enforced the learning to discover the groups of co-occurring and discriminative joints for better action recognition.

Different from these, our Part-aware LSTM (section \ref{sec:approach}) is a new RNN-based learning framework which has internal part-based memory sub-cells with a novel gating mechanism.

\begin{figure}
	\centering
	\includegraphics[width=1\linewidth]{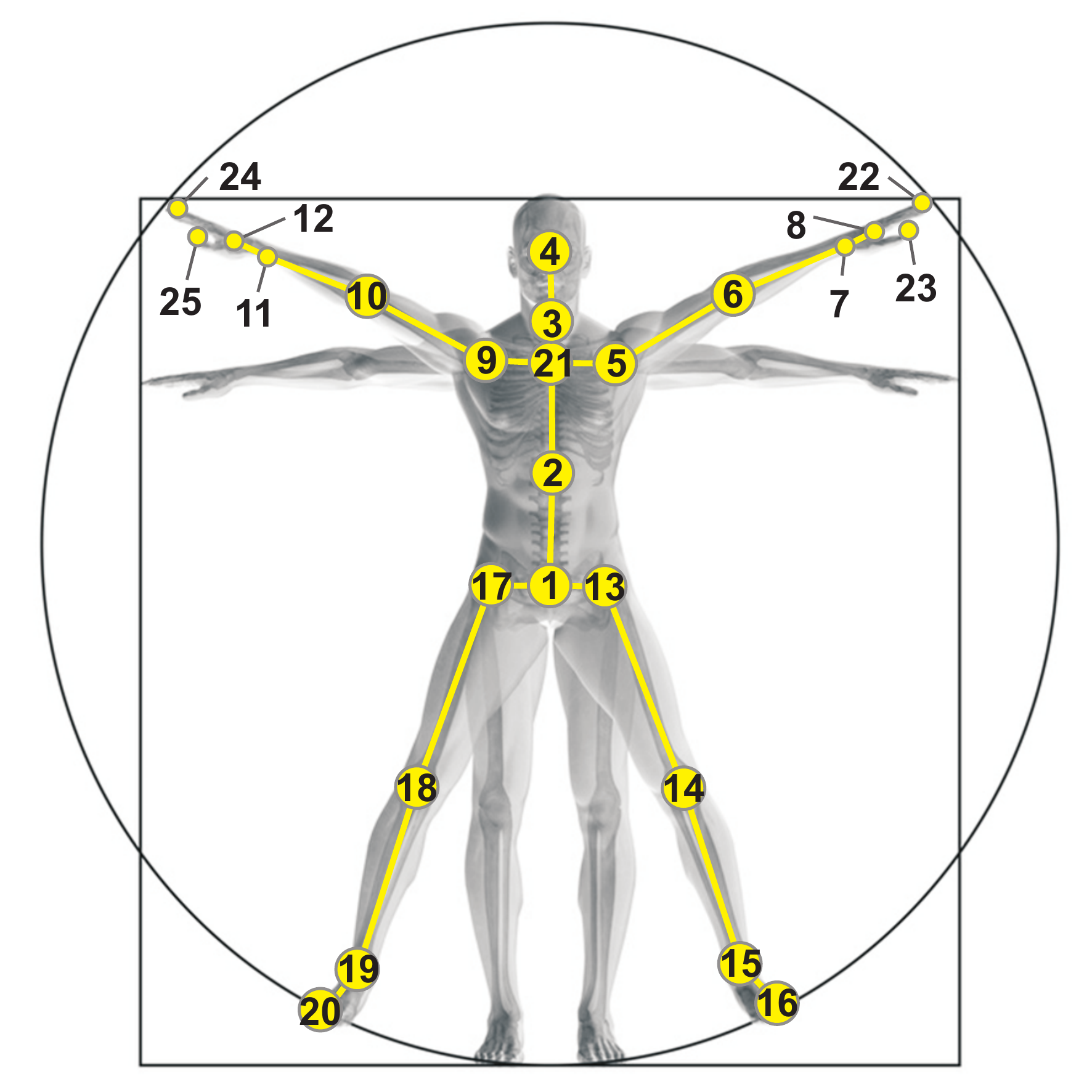}
	\caption{Configuration of 25 body joints in our dataset.
		The labels of the joints are:
		1-base~of~the~spine
		2-middle~of~the~spine
		3-neck
		4-head
		5-left~shoulder
		6-left~elbow
		7-left~wrist
		8-left~hand
		9-right~shoulder
		10-right~elbow
		11-right~wrist
		12-right~hand
		13-left~hip
		14-left~knee
		15-left~ankle
		16-left~foot
		17-right~hip
		18-right~knee
		19-right~ankle
		20-right~foot
		21-spine
		22-tip~of~the~left~hand
		23-left~thumb
		24-tip~of~the~right~hand
		25-right~thumb
	}
	\label{fig:skeleton}
\end{figure}

\section{The Dataset}
\label{sec:dataset}

This section introduces the details and the evaluation criteria of NTU RGB+D action recognition dataset.\footnote{\href{http://rose1.ntu.edu.sg/datasets/actionrecognition.asp}{http://rose1.ntu.edu.sg/datasets/actionrecognition.asp}}

\subsection{The RGB+D Action Dataset}
\label{sec:datasetstructure}

{\bf Data Modalities:} To collect this dataset, we utilized Microsoft Kinect v2 sensors.
We collected four major data modalities provided by this sensor:
depth maps, 3D joint information, RGB frames, and IR sequences.

Depth maps are sequences of two dimensional depth values in millimeters.
To maintain all the information, we applied lossless compression for each individual frame.
The resolution of each depth frame is $512\times424$.

Joint information consists of 3-dimensional locations of 25 major body joints for detected and tracked human bodies in the scene.
The corresponding pixels on RGB frames and depth maps are also provided for each joint and every frame.
The configuration of body joints is illustrated in \figurename{~\ref{fig:skeleton}}.

RGB videos are recorded in the provided resolution of $1920\times1080$.

Infrared sequences are also collected and stored frame by frame in $512\times424$.

{\bf Action Classes:} We have 60 action classes in total, which are divided into three major groups:
40 daily actions (drinking, eating, reading, etc.), 
9 health-related actions (sneezing, staggering, falling down, etc.), 
and 11 mutual actions (punching, kicking, hugging, etc.).

%
%

{\bf Subjects:} We invited 40 distinct subjects for our data collection.
The ages of the subjects are between $10$ and $35$.
\figurename{~\ref{fig:sampleframes}} shows the variety of the subjects in age, gender, and height.
Each subject is assigned a consistent ID number over the entire dataset.

{\bf Views:} We used three cameras at the same time to capture three different horizontal views from the same action.
For each setup, the three cameras were located at the same height but from three different horizontal angles: $-45^\circ, 0^\circ, +45^\circ$. Each subject was asked to perform each action twice, once towards the left camera and once towards the right camera.
In this way, we capture two front views, one left side view, one right side view, one left side 45 degrees view, and one right side 45 degrees view.
The three cameras are assigned consistent camera numbers.
Camera~1 always observes the 45 degrees views, while camera 2 and 3 observe front and side views.

To further increase the camera views, on each setup we changed the height and distances of the cameras to the subjects, as reported in \tablename{~\ref{tab:camerasetups}}.
All the camera and setup numbers are provided for each video sample.

\subsection{Benchmark Evaluations}
\label{sec:evaluations}

To have standard evaluations for all the reported results on this benchmark, we define precise criteria for two types of action classification evaluation, as described in this section.
For each of these two, we report the classification accuracy in percentage.

\subsubsection{Cross-Subject Evaluation}
\label{sec:crosssubjevaluation}
In cross-subject evaluation, we split the 40 subjects into training and testing groups.
Each group consists of 20 subjects.
For this evaluation, the training and testing sets have $40,320$ and $16,560$ samples, respectively.
The IDs of training subjects in this evaluation are:
1, 2, 4, 5, 8, 9, 13, 14, 15, 16, 17, 18, 19, 25, 27, 28, 31, 34, 35, 38; 
remaining subjects are reserved for testing.

\subsubsection{Cross-View Evaluation}
\label{sec:crosssubjevaluation}
For cross-view evaluation, we pick all the samples of camera 1 for testing and samples of cameras 2 and 3 for training.
In other words, the training set consists of front and two side views of the actions, while testing set includes left and right 45 degree views of the action performances.
For this evaluation, the training and testing sets have $37,920$ and $18,960$ samples, respectively.

\begin{table}
	\setlength{\tabcolsep}{4pt} 
	\begin{center}
		\begin{tabular}{|c||c|c||c||c|c|}
			\hline
			Setup  & Height & Distance	& Setup  & Height & Distance \\
			No.  & (m) & (m) & No.  & (m) & (m) \\\hline\hline
			1  & 1.7 & 3.5 & 2  & 1.7 & 2.5\\\hline
			3  & 1.4 & 2.5 & 4  & 1.2 & 3.0\\\hline
			5  & 1.2 & 3.0 & 6  & 0.8 & 3.5\\\hline
			7  & 0.5 & 4.5 & 8  & 1.4 & 3.5\\\hline
			9  & 0.8 & 2.0 & 10 & 1.8 & 3.0\\\hline
			11 & 1.9 & 3.0 & 12 & 2.0 & 3.0\\\hline
			13 & 2.1 & 3.0 & 14 & 2.2 & 3.0\\\hline
			15 & 2.3 & 3.5 & 16 & 2.7 & 3.5\\\hline
			17 & 2.5 & 3.0 & ~  &  ~  &  ~ \\\hline
		\end{tabular}
	\end{center}
	\caption{Height and distance of the three cameras for each collection setup.
		All height and distance values are in meters.}
	\label{tab:camerasetups}
\end{table}

\section{Part-Aware LSTM Network}
\label{sec:approach}

In this section, we introduce a new data-driven learning method to model the human actions using our collected 3D action sequences.

Human actions can be interpreted as time series of body configurations.
These body configurations can be effectively and succinctly represented by the 3D locations of major joints of the body.
In this fashion, each video sample can be modeled as a sequential representation of configurations.

Recurrent Neural Networks (RNNs) and Long Short-Term Memory Networks (LSTMs) \cite{lstm} have been shown to be among the most successful deep learning models to encode and learn sequential data in various applications \cite{seq2seq,Donahue_2015_CVPR,Byeon_2015_CVPR,char-rnn-arxiv}.

In this section, we introduce the traditional recurrent neural networks and then propose our part-aware LSTM model.

\subsection{Traditional RNN and LSTM}
\label{sec:rnnlstm}

A recurrent neural network transforms an input sequence $(\bf X)$ to another sequence $(\bf Y)$ by updating its internal state representation $({\bf h}_t)$ at each time step $(t)$ as a linear function of the last step's state and the input at the current step, followed by a nonlinear scaling function.
Mathematically:
\begin{eqnarray}
{\bf h}_t &=& \sigma\begin{pmatrix}{\bf W} \begin{pmatrix} {\bf x}_t \\ {\bf h}_{t-1} \end{pmatrix}\end{pmatrix}\\
{\bf y}_t &=& \sigma\begin{pmatrix}{\bf V} {\bf h}_t\end{pmatrix}
\end{eqnarray}
where $t\in \{1,..,T\}$ represents time steps, and $\sigma\in\{Sigm, Tanh \}$ is a nonlinear scaling function.

Layers of RNNs can be stacked to build a deep recurrent network:

\begin{eqnarray}
{\bf h}_t^l &=& \sigma\begin{pmatrix}{\bf W}^l \begin{pmatrix} {\bf h}_t^{l-1} \\ {\bf h}_{t-1}^l \end{pmatrix}\end{pmatrix}\\
{\bf h}_t^0 &:=& {\bf x}_t\\
{\bf y}_t &=& \sigma\begin{pmatrix}{\bf V} {\bf h}_t^L\end{pmatrix}
\end{eqnarray}
where $l\in\{1,...,L\}$ represents layers.

\begin{figure}
	\centering
	\def\svgwidth{\columnwidth}
	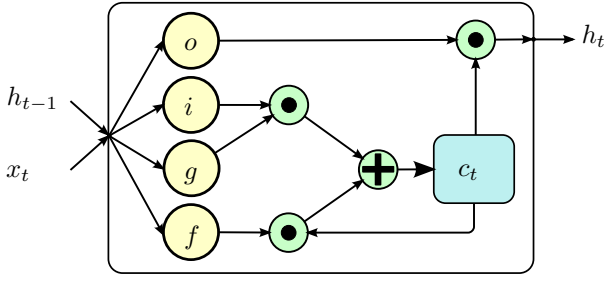
	\caption{Schema of a long short-term memory (LSTM) unit.
		$o$ is the output gate,
		$i$ is the input gate,
		$g$ is the input modulation gate,
		and $f$ is the forget gate.
		$c$ is the memory cell to keep the long term context.
	}
	\label{fig:lstm}
\end{figure}

Traditional RNNs have limited abilities to keep long-term representation of the sequences and were unable to discover relations among long-ranges of inputs.
To alleviate this drawback, Long Short-Term Memory Network \cite{lstm} was introduced to keep a long term memory inside each RNN unit and learn when to remember or forget information stored inside its internal memory cell $(c^t)$:

\begin{eqnarray}
\begin{pmatrix}i\\f\\o\\g\end{pmatrix} &=& \begin{pmatrix}Sigm\\Sigm\\Sigm\\Tanh\end{pmatrix}
\begin{pmatrix}{\bf W}\begin{pmatrix} {\bf x}_t \\ {\bf h}_{t-1} \end{pmatrix}\end{pmatrix}\\
c_t&=&f\odot c_{t-1} + i\odot g\\
h_t&=&o\odot Tanh(c_t)
\end{eqnarray}
In this model, $i, f, o$, and $g$ denote input gate, forget gate, output gate, and input modulation gate respectively.
Operator $\odot$ denotes element-wise multiplication.
\figurename{~\ref{fig:lstm}} shows the schema of this recurrent unit.

The output ${\bf y}_t$ is fed to a softmax layer to transform the output codes to probability values of class labels.
To train such networks for action recognition, we fix the training output label for each input sample over time.

\subsection{Proposed Part-Aware LSTM}
\label{sec:plstm}

In human actions, body joints move together in groups.
Each group can be assigned to a major part of the body, and actions can be interpreted based on the interactions between body parts or with other objects.
Based on this intuition, we propose a part-aware LSTM human action learning model.
We dub the method P-LSTM.

Instead of keeping a long-term memory of the entire body's motion in the cell, we split it to part-based cells.
It is intuitive and more efficient to keep the context of each body part independently and represent the output of the P-LSTM unit as a combination of independent body part context information.
In this fashion, each part's cell has its individual input, forget, and modulation gates, but the output gate will be shared among the body parts.
In our model, we group the body joints into five part groups: torso, two hands, and two legs.

At each frame $t$, we concatenate the 3D coordinates of the joints inside each part $p\in\{1,...,P\}$ and consider them as the input representation of that part, denoted as ${\bf x}_t^p$.

Thusly, the proposed P-LSTM is modeled as:
\begin{eqnarray}
\begin{pmatrix}i^p\\f^p\\g^p\end{pmatrix} &=& \begin{pmatrix}Sigm\\Sigm\\Tanh\end{pmatrix}
\begin{pmatrix}{\bf W}^p \begin{pmatrix} {\bf x}_t^p \\ {\bf h}_{t-1}\end{pmatrix}\end{pmatrix}\\
c_t^p&=&f^p\odot c_{t-1}^p + i^p\odot g^p\\
o &=& Sigm \begin{pmatrix}{\bf W}_o \begin{pmatrix} {\bf x}_t^1 \\ \vdots \\{\bf x}_t^P \\{\bf h}_{t-1} \end{pmatrix}\end{pmatrix}\\
h_t&=&o\odot Tanh\begin{pmatrix}{\bf c}_t^1 \\ \vdots \\{\bf c}_t^P\end{pmatrix}
\end{eqnarray}

\begin{figure}
	\centering
	\def\svgwidth{\columnwidth}
	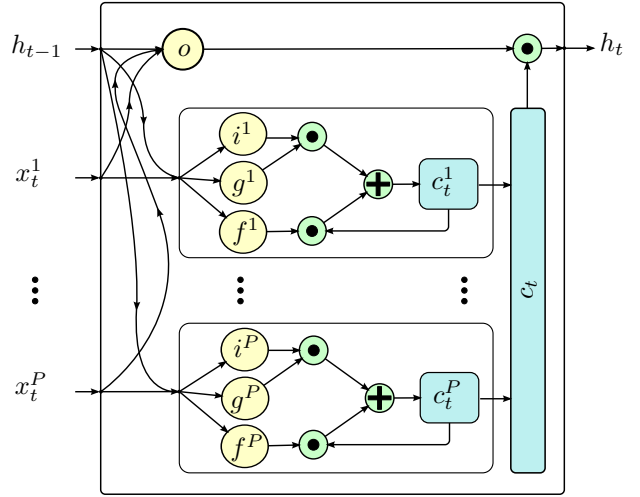
	\caption{Illustration of the proposed part-aware long short-term memory (P-LSTM) unit.}
	\label{fig:plstm}
\end{figure}

A graphical representation of the propsed P-LSTM is illustrated in \figurename{~\ref{fig:plstm}}.

The LSTM baseline has full connections between all the memory cells and all the input features via input modulation gate and the memory cell was supposed to represent the long-term dynamics of the entire skeleton over time.
This leads to a very large size of training parameters which are prone to overfitting.
We propose to regularize this by dropping unnecessary links.
We divide the entire body's dynamics (represented in the memory cell) to the dynamics of body parts (part-based cells) and learn the final classifier over their concatenation.
Our P-LSTM learns the common temporal patterns of the parts independently and combines them in the global level representation for action recognition.

\section{Experiments}
\label{sec:exp}

In our experiments, we evaluate state-of-the-art depth-based action recognition methods and compare them with RNN, LSTM, and the proposed P-LSTM based on the evaluation criteria of our dataset.

\subsection{Experimental Setup}
\label{sec:expsetup}

We use the publicly available implementation of six depth-based action recognition methods and apply them on our new dataset benchmark.
Among them, HOG$^2$ \cite{hog2-ohnbar}, Super Normal Vector \cite{Yang_2014_CVPR}, and HON4D \cite{HON4D} extract features directly from depth maps without using the skeletal information.
Lie group \cite{VemulapalliCVPR14}, Skeletal Quads \cite{skeletalQuads}, and FTP Dynamic Skeletons \cite{jianfang_CVPR15} are skeleton-based methods.

The other evaluated methods are RNN, LSTM, and the proposed P-LSTM method.

For skeletal representation, we apply a normalization preprocessing step.
The original 3D locations of the body joints are provided in camera coordinate system.
We translate them to the body coordinate system with its origin on the ``middle of the spine'' joint (number 2 in \figurename{~\ref{fig:skeleton}}), followed by a 3D rotation to fix the $X$ axis parallel to the 3D vector from ``right shoulder'' to ``left shoulder'', and $Y$ axis towards the 3D vector from ``spine base'' to ``spine''. 
The $Z$ axis is fixed as the new $X\times Y$.
In the last step of normalization, we scale all the 3D points based on the distance between ``spine base'' and ``spine'' joints.

In the cases of having more than one body in the scene, we transform all of them with regard to the main actor's skeleton.
To choose the main actor among the available skeletons, we pick the one with the highest amount of 3D body motion.

Kinect's body tracker is prone to detecting some objects \eg seats or tables as bodies.
To filter out these noisy detections, for each tracked skeleton we calculate the spread of the joint locations towards image axis and filtered out the ones whose $X$ spread were more than $0.8$ of their $Y$ spread.

For our recurrent model evaluation, we reserve about five percent of the training data as validation set.
The networks are trained on a large number of iterations and we pick the network with the least validation error among all the iterations and report its performance on testing data.

For each video sample at each training iteration, we split the video to $T=8$ equal sized temporal segments and randomly pick one frame from each segment to feed the skeletal information of that frame as input to the recurrent leaning models in $t\in \{1,...,T\}$ time steps.

For the baseline methods which use SVM as their classifier, to be able to manage the large scale of the data, we use Libliner SVM toolbox \cite{liblinear}.

Our RNN, LSTM, and P-LSTM implementations are done on the Torch toolbox platform \cite{torch}.
We use a Nvidia Tesla K40 GPU to run our experiments.

\begin{table}
	\begin{center}
		\begin{tabular}{|l|c|c|}
			\hline
			~  & Cross & Cross\\
			Method  & Subject & View \\
			~  & Accuracy & Accuracy \\
			\hline\hline
			HOG$^2$ \cite{hog2-ohnbar}					& 32.24\% & 22.27\%	\\\hline
			Super Normal Vector \cite{Yang_2014_CVPR}	& 31.82\% & 13.61\%	\\\hline
			HON4D \cite{HON4D} 							& 30.56\% & 7.26\%	\\\hline
			
			Lie Group \cite{VemulapalliCVPR14}			& 50.08\% & 52.76\%	\\\hline
			Skeletal Quads \cite{skeletalQuads}			& 38.62\% & 41.36\%	\\\hline
			FTP Dynamic Skeletons \cite{jianfang_CVPR15}& 60.23\% & 65.22\%	\\\hline
			HBRNN-L \cite{rnnskeleton_cvpr15}			& 59.07\% & 63.97\%	\\\hline
			\hline			
			1 Layer RNN 						& 56.02\% & 60.24\% \\\hline
			2 Layer RNN 						& 56.29\% & 64.09\% \\\hline
			\hline
			1 Layer LSTM 						& 59.14\% & 66.81\% \\\hline
			2 Layer LSTM 						& 60.69\% & 67.29\%	\\\hline
			\hline
			1 Layer P-LSTM 						& 62.05\% & 69.40\%	\\\hline
			\bf 2 Layer P-LSTM 					& \bf 62.93\% & \bf 70.27\%	\\\hline
		\end{tabular}
	\end{center}
	\caption{The results of the two evaluation settings of our benchmark using different methods.
		First three rows are depth-map based baseline methods.
		Rows 4, 5, and 6 are three skeleton-based baseline methods.
		Following rows report the performance of RNN, LSTM and the proposed P-LSTM model.
		Our P-LSTM learning model outperforms other methods on both of the evaluation settings.}
	\label{tab:results}
\end{table}

\subsection{Experimental Evaluations}
\label{sec:expeval}

The results of our evaluations of the above-mentioned methods are reported in \tablename{~\ref{tab:results}}.
First three rows show the accuracies of the evaluated depth-map features.
They perform better in cross-subject evaluation compared to the cross-view one.
The reason for this difference is that in the cross-view scenario, the depth appearance of the actions are different and these methods are more prone to learning the appearances or view-dependent motion patterns.

Skeletal-based features (Lie group \cite{VemulapalliCVPR14}, Skeletal Quads \cite{skeletalQuads}, and FTP Dynamic Skeletons \cite{jianfang_CVPR15}), perform better with a notable gap on both settings.
They are stronger to generalize between the views because the 3D skeletal representation is view-invariant in essence, but it's prone to errors of the body tracker.

As the most relevant baseline, we implemented HBRNN-L \cite{rnnskeleton_cvpr15} which achieved competitive results to the best hand-crafted methods.
Although \cite{rnnskeleton_cvpr15} reported the ineffectiveness of dropout on their experiments, we found it effective on all of our evaluations (including their method).
This shows they have their model was prone to overfitting due to the lack of training data and proves the demand for a bigger dataset and approves our motivation for proposing NTU RGB+D dataset.

At the next step, we evaluate the discussed recurrent networks on this benchmark.
Although RNN has the limitation in discovering long-term interdependency of inputs, they perform competitively with the hand-crafted methods.
Stacking one more RNN layer improves the overall performance of the network, especially in cross-view scenario.


By utilizing long-term context in LSTM, the performances are improved significantly.
LSTM's performance improves slightly by stacking one more layer.

At the last step, we evaluate the proposed P-LSTM model.
By isolating the context memory of each body part and training the classifier based on their combination, we model a new way of regularization in the learning process of LSTM parameters.
It utilizes the high intra-part and low inter-part correlation of input features to improve the learning process of the LSTM network.
As shown in \tablename{~\ref{tab:results}} P-LSTM outperforms all other methods by achieving 62.93\% in cross-subject, and 70.27\% in cross-view evaluations.

\begin{figure*}
	\centering
	\setlength{\tabcolsep}{2pt} 
	\begin{tabular}{ccccc}		
		\includegraphics[width=84pt]{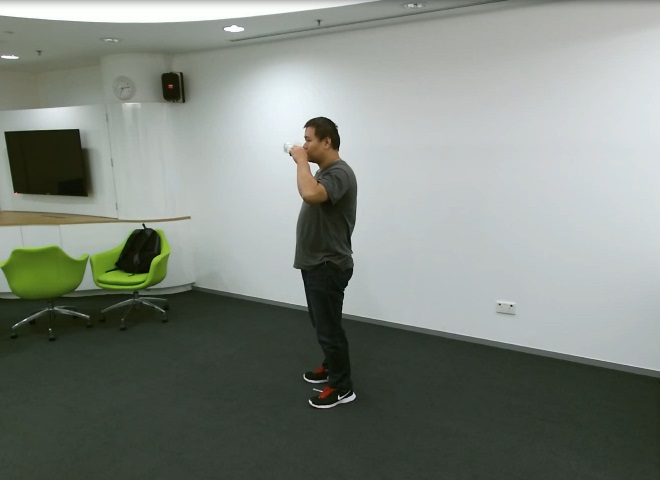} & 
		\includegraphics[width=84pt]{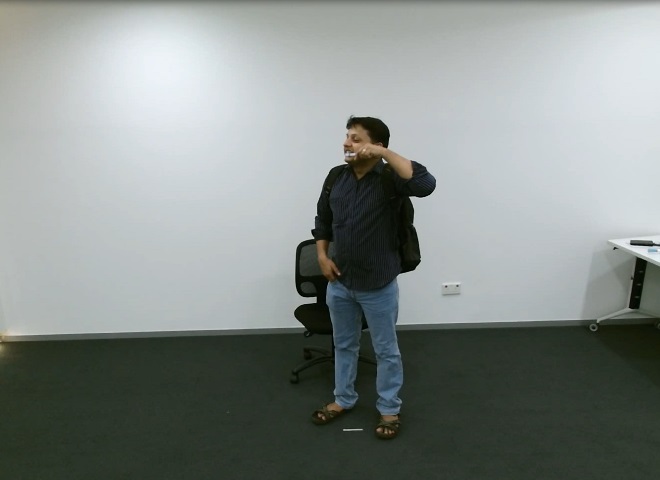} & 
		\includegraphics[width=84pt]{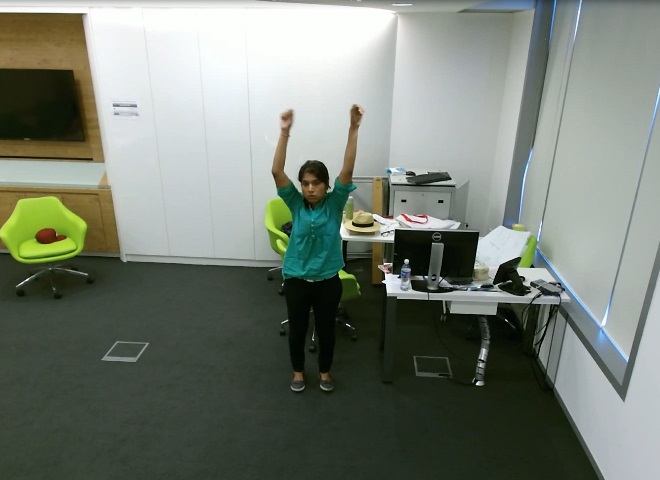} & 
		\includegraphics[width=84pt]{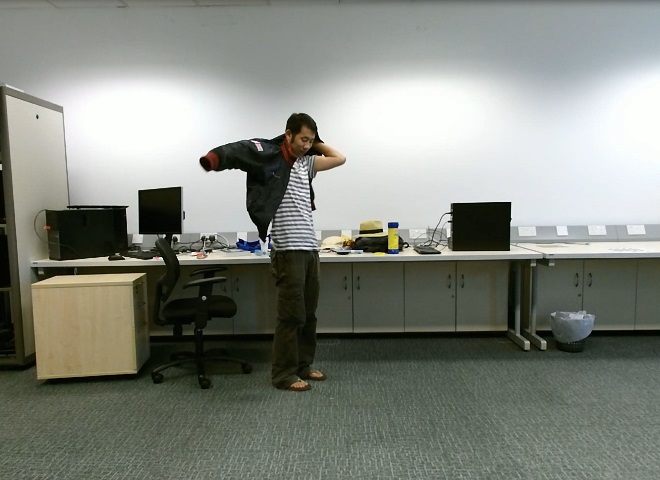} & 
		\includegraphics[width=84pt]{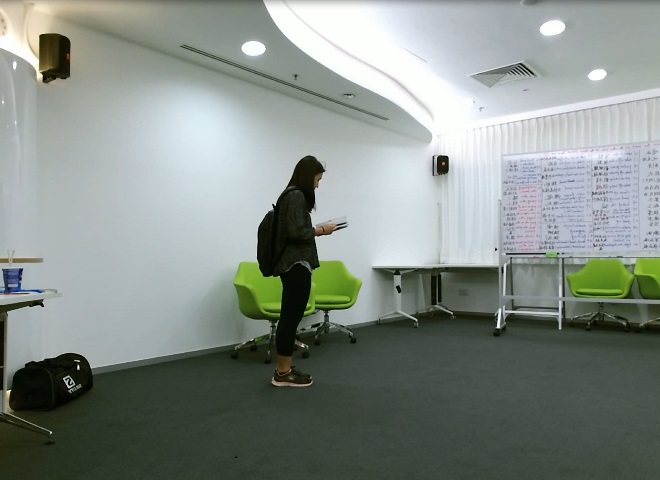} \\
		\includegraphics[width=84pt]{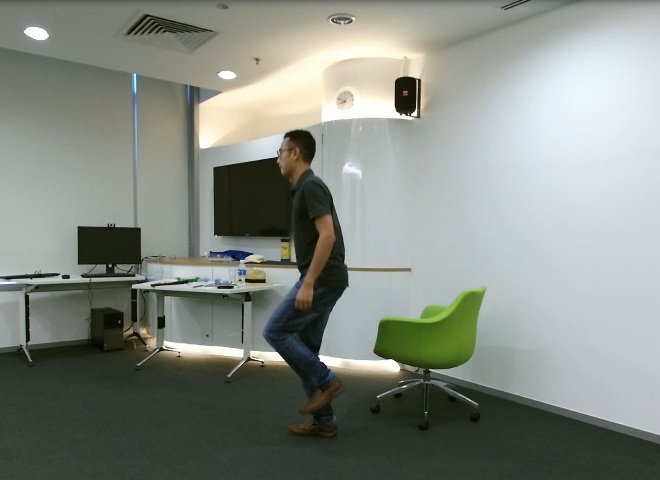} & 
		\includegraphics[width=84pt]{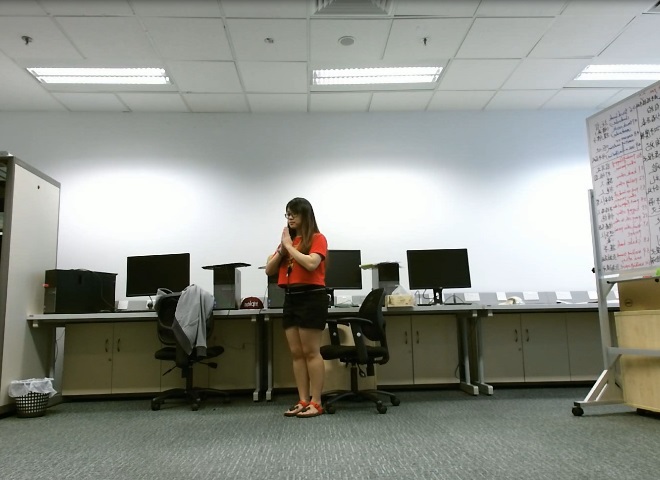} & 
		\includegraphics[width=84pt]{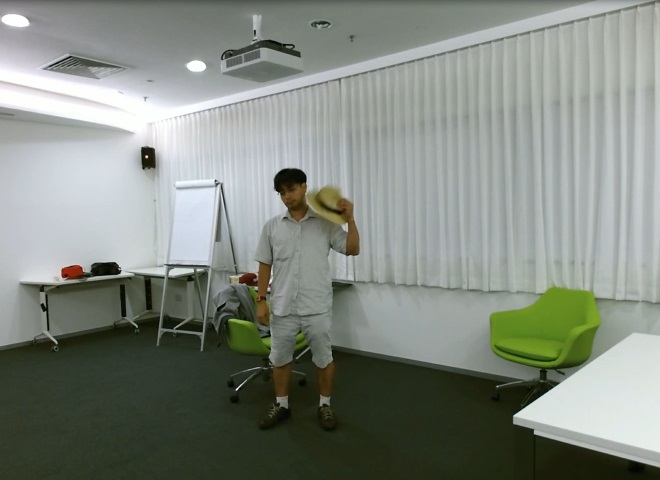} & 
		\includegraphics[width=84pt]{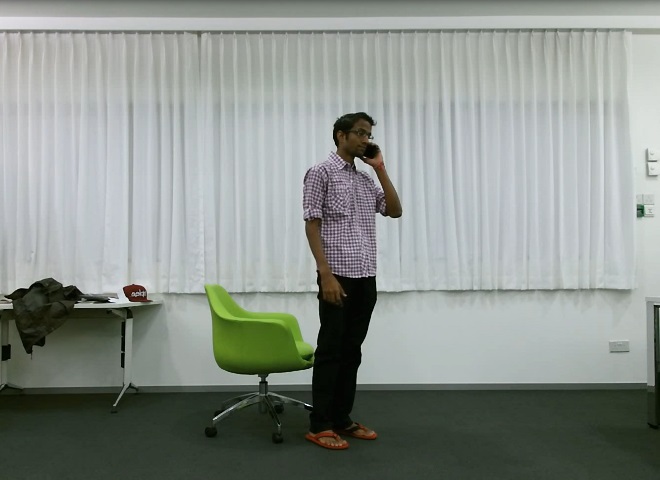} & 
		\includegraphics[width=84pt]{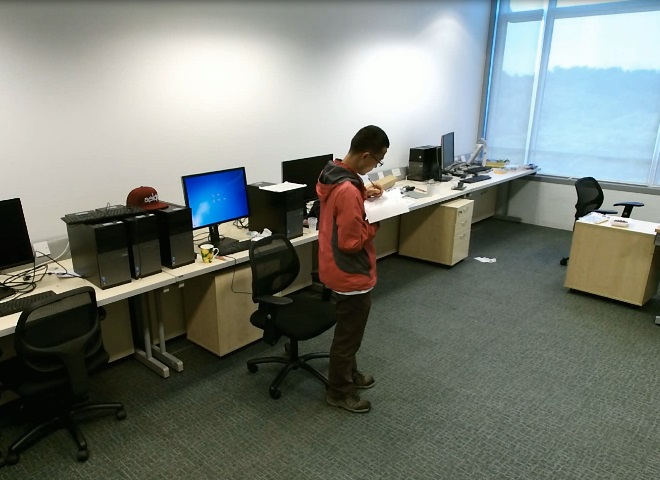} \\
		\includegraphics[width=84pt]{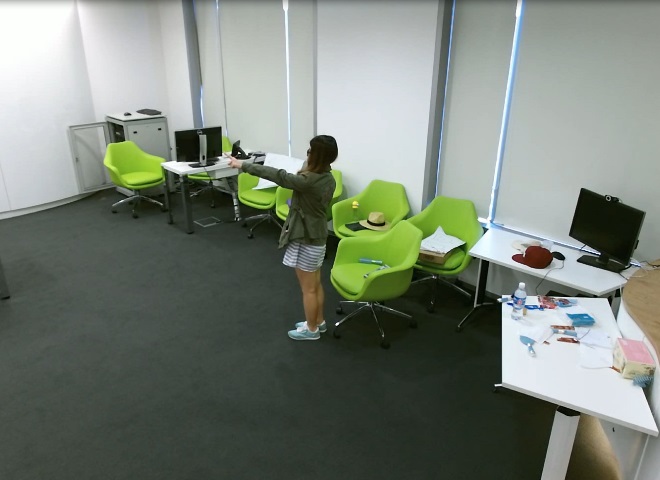} & 
		\includegraphics[width=84pt]{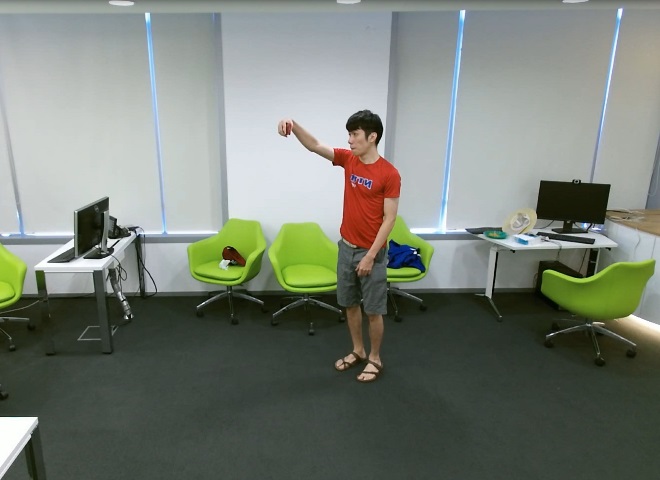} & 
		\includegraphics[width=84pt]{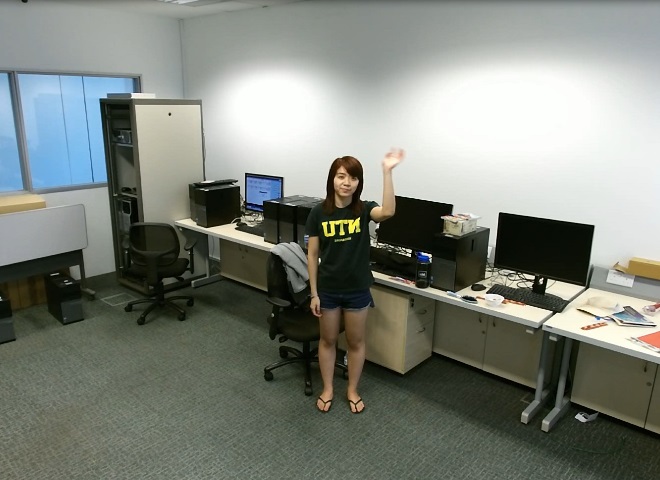} & 
		\includegraphics[width=84pt]{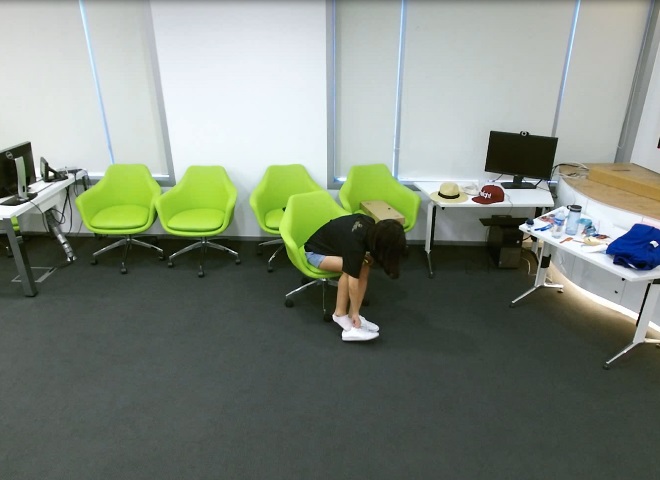} & 
		\includegraphics[width=84pt]{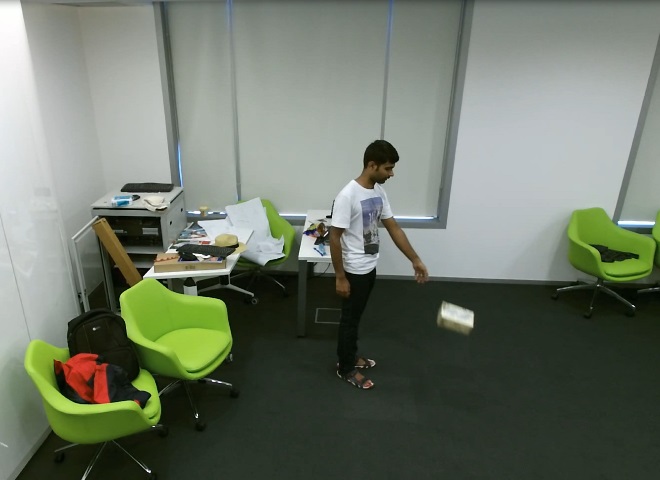} \\
		\includegraphics[width=84pt]{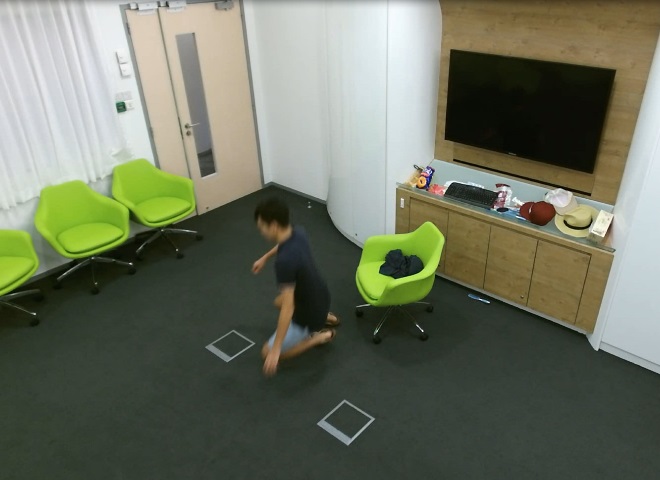} & 
		\includegraphics[width=84pt]{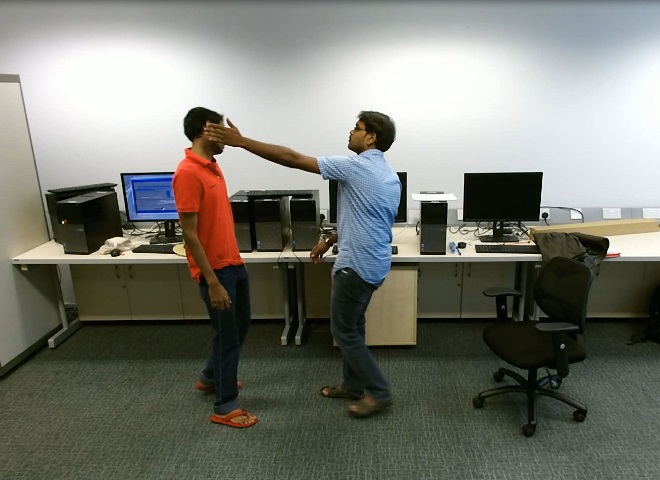} & 
		\includegraphics[width=84pt]{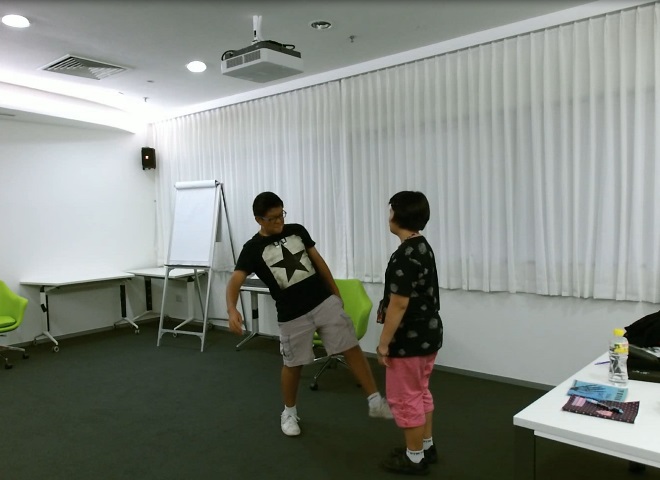} & 
		\includegraphics[width=84pt]{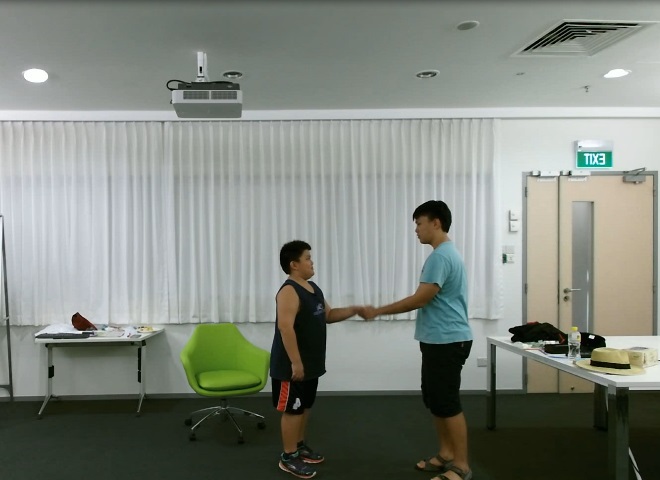} & 
		\includegraphics[width=84pt]{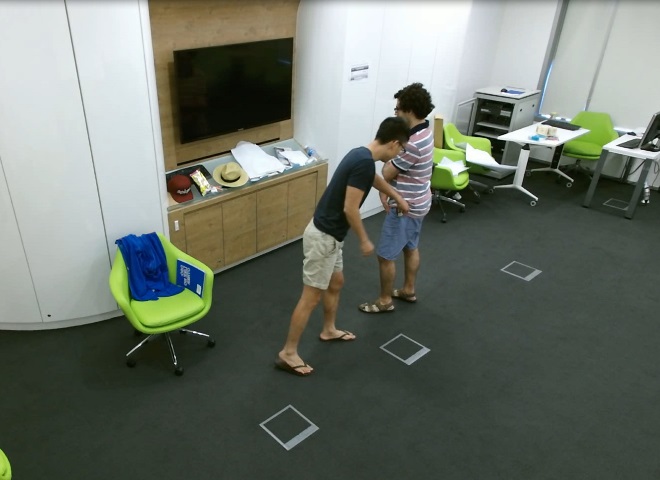} \\&&&&\\
		\includegraphics[width=84pt]{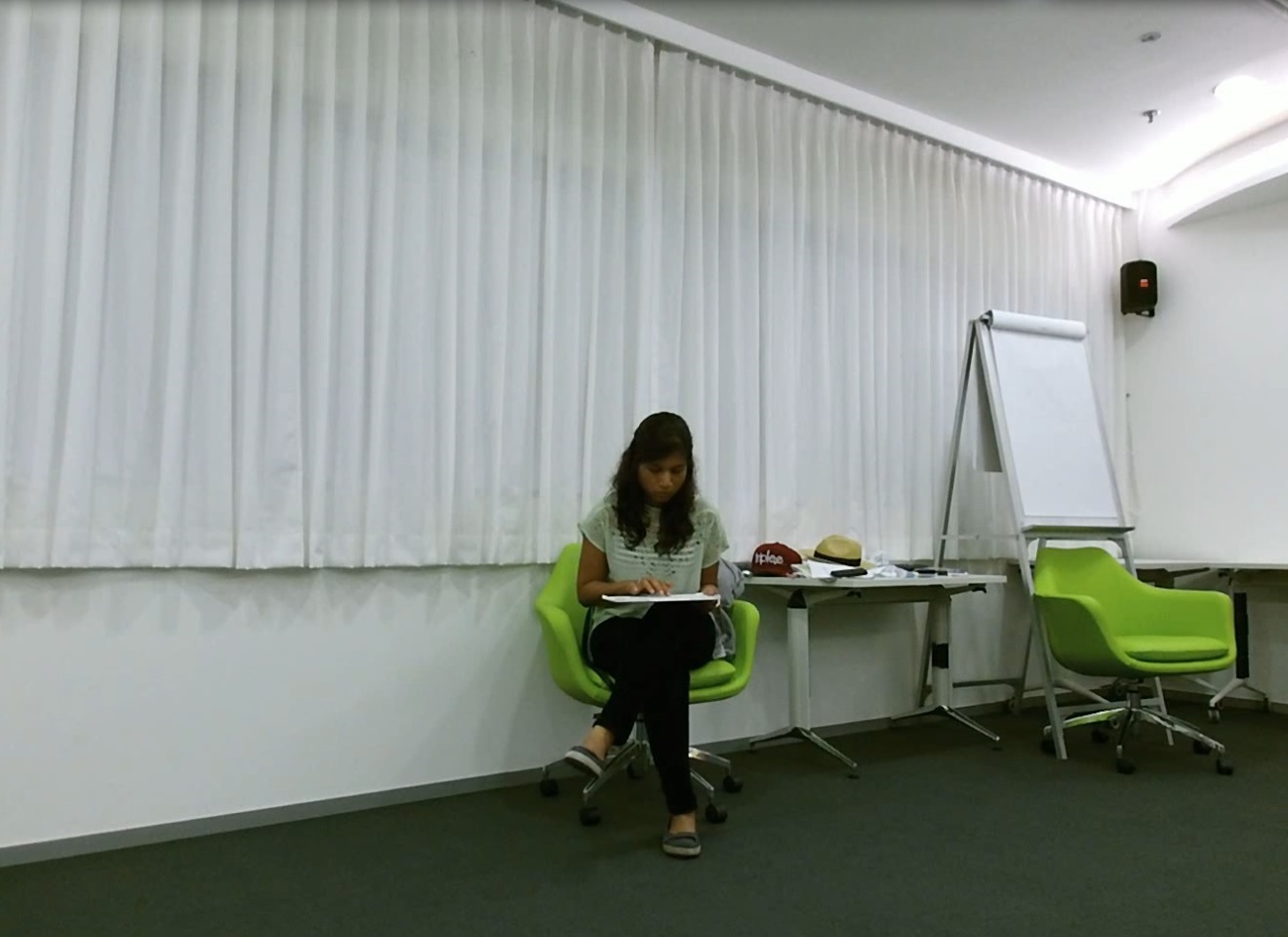} & 
		\includegraphics[width=84pt]{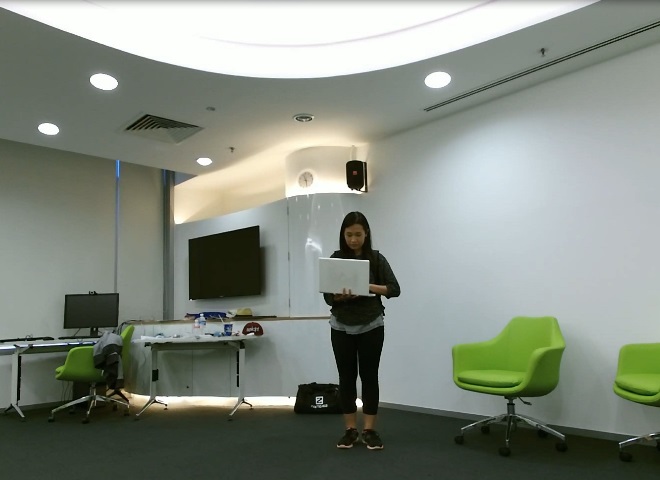} & 
		\includegraphics[width=84pt]{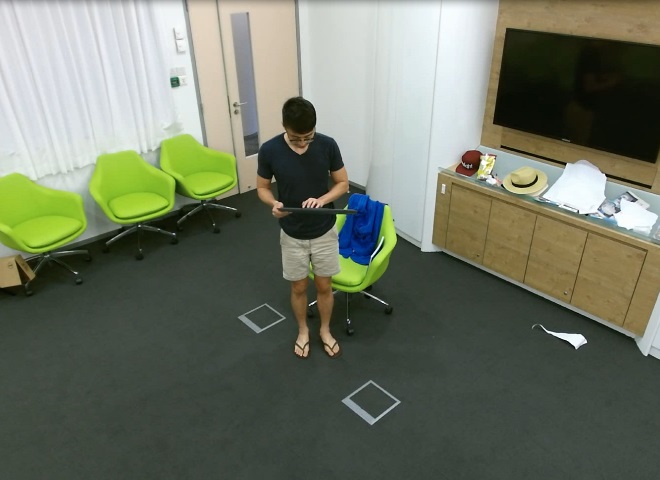} & 
		\includegraphics[width=84pt]{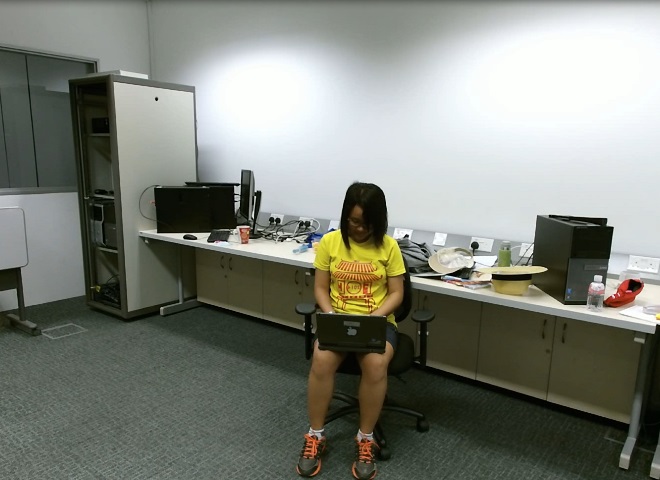} & 
		\includegraphics[width=84pt]{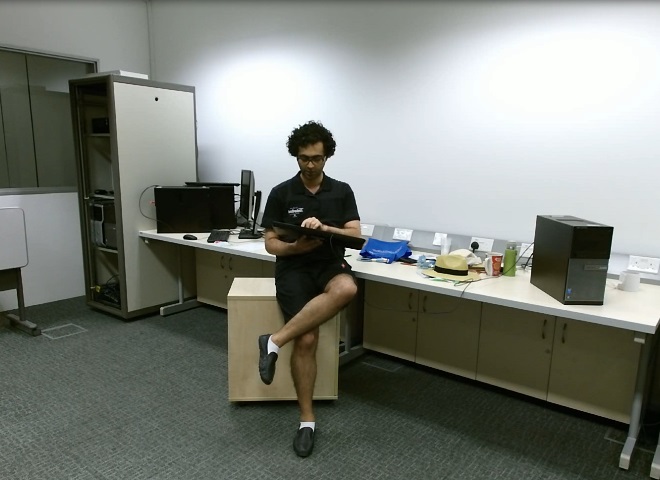} \\
		\includegraphics[width=84pt]{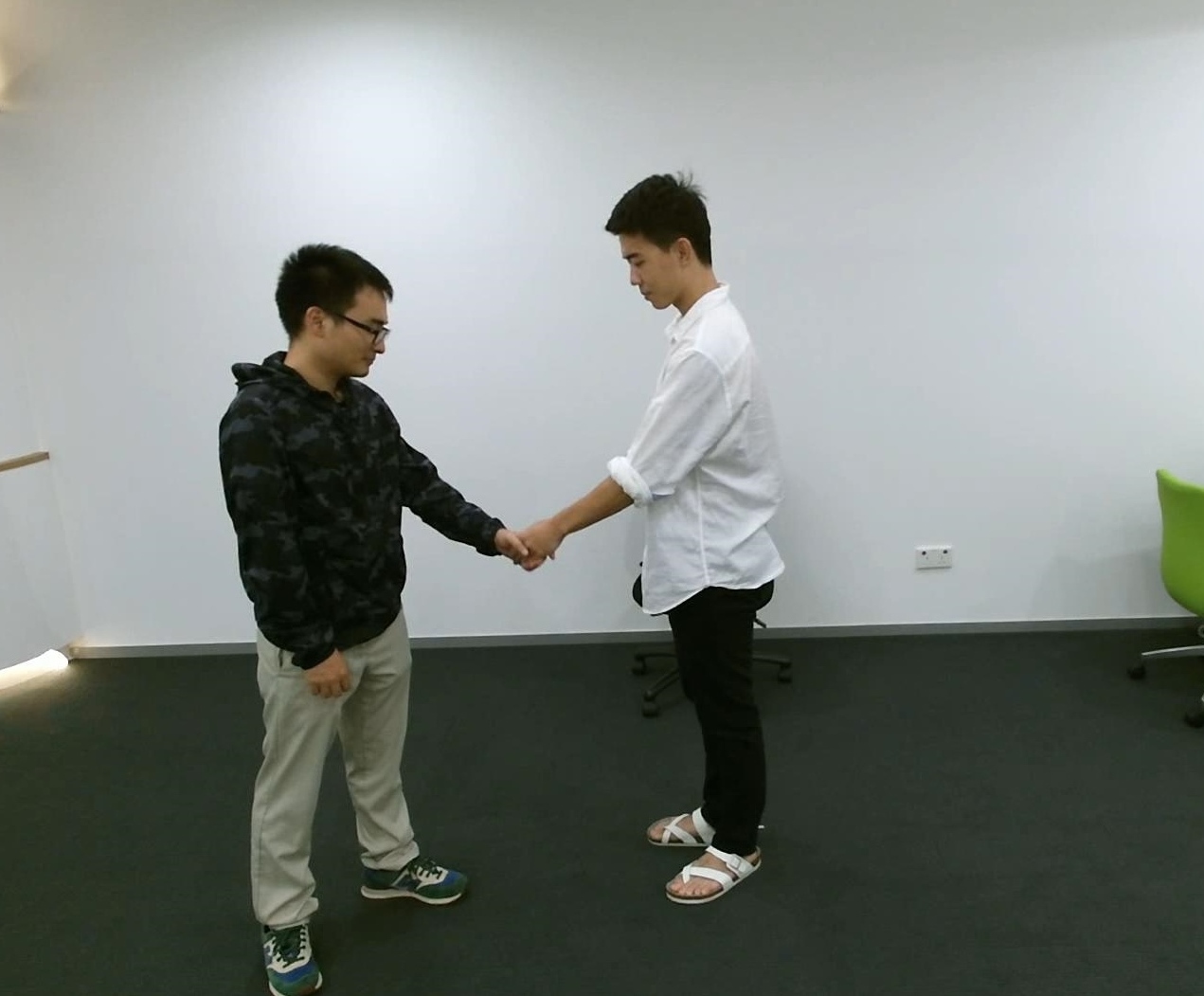} & 
		\includegraphics[width=84pt]{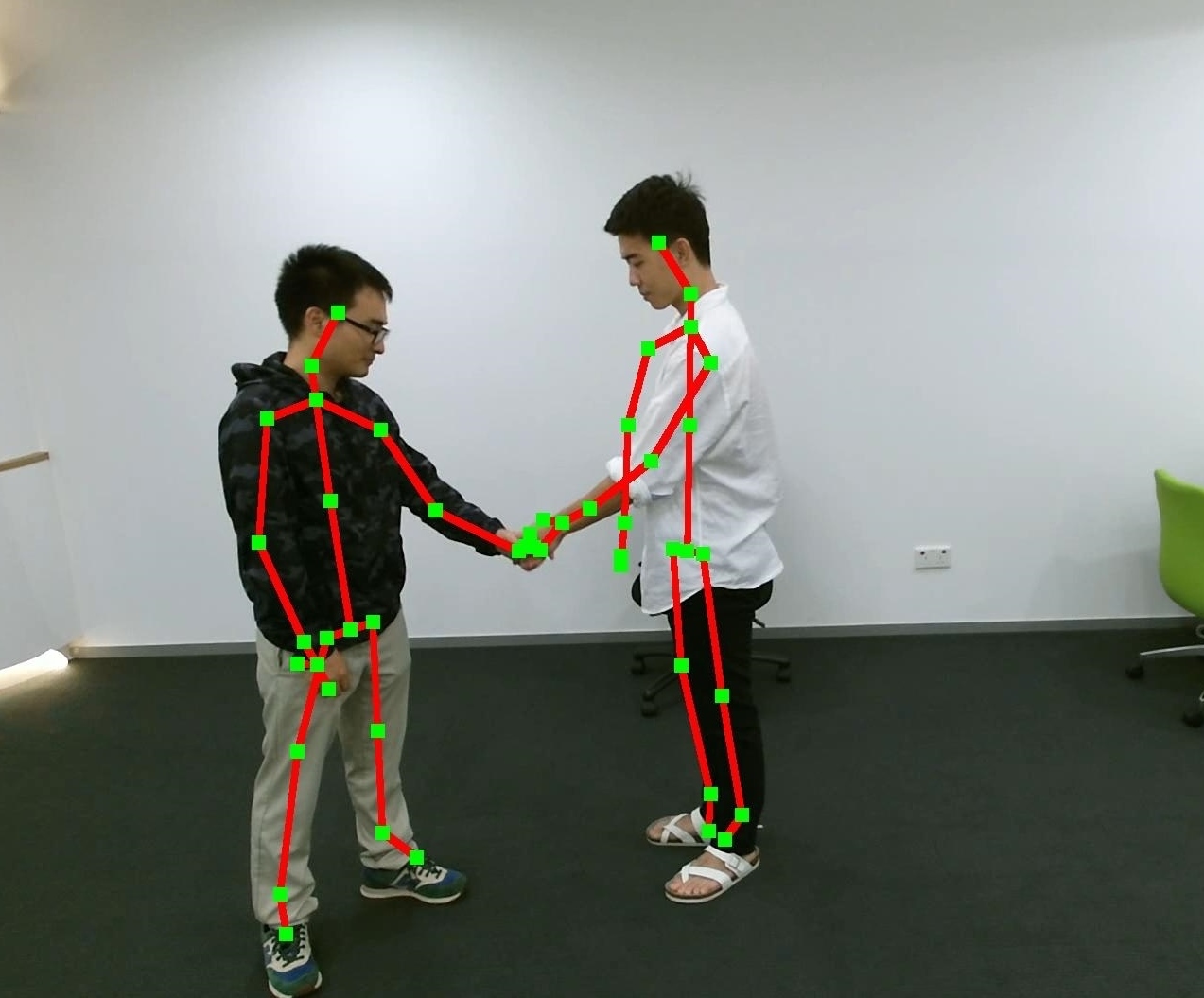} & 
		\includegraphics[width=84pt]{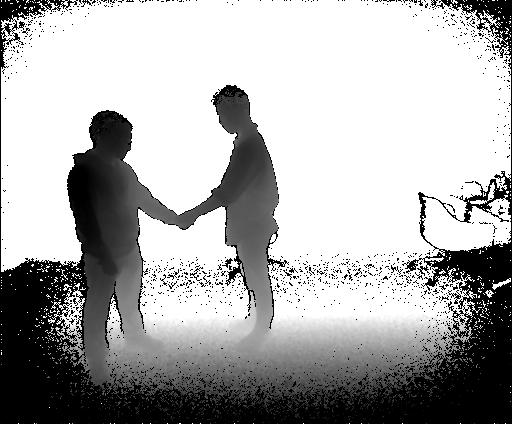} & 
		\includegraphics[width=84pt]{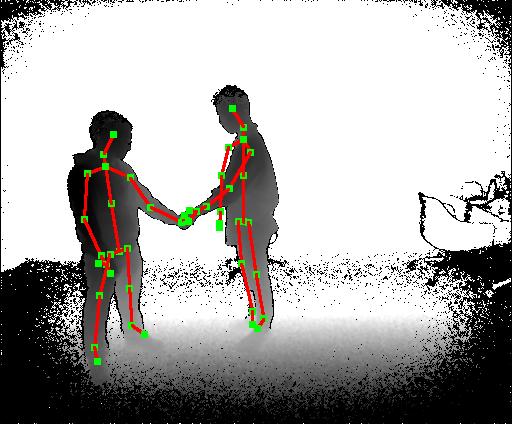} & 
		\includegraphics[width=84pt]{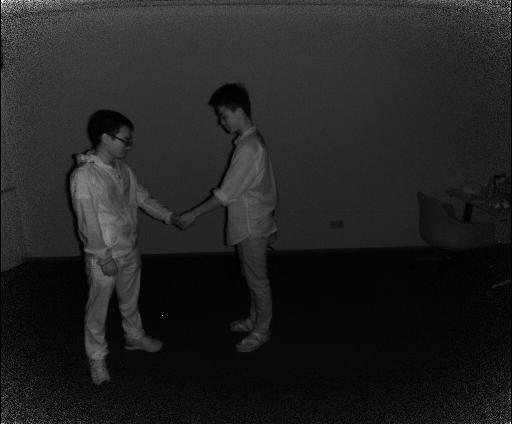} \\
	\end{tabular}
	\vspace{2pt}
	\caption{Sample frames of the NTU RGB+D dataset.
		First four rows show the variety in human subjects and camera views.
		Fifth row depicts the intra-class variation of the performances.
		The last row illustrates RGB, RGB+joints, depth, depth+joints, and IR modalities of a sample frame.}
	\label{fig:sampleframes}
\end{figure*}

\section{Conclusion}
\label{sec:conclusion}

A large-scale RGB+D action recognition dataset is introduced in this paper.
Our dataset includes 56880 video samples collected from 60 action classes in highly variant camera settings.
Compared to the current datasets for this task, our dataset is larger in orders and contains much more variety in different aspects.

The large scale of the collected data enables us to apply data-driven learning methods like Long Short-Term Memory networks in this problem and achieve better performance accuracies compared to hand-crafted features.

We also propose a Part-aware LSTM model to utilize the physical structure of the human body to further improve the performance of the LSTM learning framework.

The provided experimental results show the availability of large-scale data enables the data-driven learning frameworks to outperform hand-crafted features.
They also show the effectiveness of the proposed P-LSTM model over traditional recurrent models.

\section{Acknowledgement}
This research was carried out at the Rapid-Rich Object Search (ROSE) Lab at the Nanyang Technological University, Singapore.  The ROSE Lab is supported by the National Research Foundation, Singapore, under its Interactive Digital Media (IDM) Strategic Research Programme.

The research is in part supported by Singapore Ministry of Education (MOE) Tier 2 ARC28/14, and Singapore A*STAR Science and Engineering Research Council PSF1321202099.

We gratefully acknowledge the support of NVIDIA Corporation with the donation of the Tesla K40 GPU used for this research.


{\small
	\bibliographystyle{ieee}
	\bibliography{D:/Users/Amir/Dropbox/Literature/AmirPhdCon}
}

\end{document}